\documentclass[utf8]{FrontiersinHarvard} % for articles in journals using the Harvard Referencing Style (Author-Date), for Frontiers Reference Styles by Journal: https://zendesk.frontiersin.org/hc/en-us/articles/360017860337-Frontiers-Reference-Styles-by-Journal  
\usepackage{url,hyperref,lineno,subcaption}
\usepackage[nopatch=footnote]{microtype}
\usepackage[onehalfspacing]{setspace}

\usepackage{bbm}
\usepackage{booktabs}
\usepackage{multirow}
\usepackage{makecell}
\usepackage{amsmath,amssymb}
\usepackage{mathtools}

\newcommand{\vect}[1]{\mathbf{#1}}

\newcommand{\set}[1]{\mathcal{#1}}

\DeclareMathOperator{\simop}{sim}

\def\keyFont{\fontsize{8}{11}\helveticabold }
\def\firstAuthorLast{Zhang {et~al.}} %use et al only if is more than 1 author
\def\Authors{
	Xianqi Zhang\,$^{1, 3}$, 
	Xingtao Wang\,$^{1, *}$ 
	and 
	Xiaopeng Fan\,$^{1, 2, 3}$
}
\def\Address{
	$^{1}$Faculty of Computing, Harbin Institute of Technology, Harbin 150001, China \\
	$^{2}$Peng Cheng Laboratory, Shenzhen 518055, China \\
	$^{3}$Suzhou Research Institute, Harbin Institute of Technology, Suzhou 215104, China
}
\def\corrAuthor{Corresponding Author}
\def\corrEmail{xtwang@hit.edu.cn}

\begin{document}
	\onecolumn
	\firstpage{1}
	
	\title[PHARL]{
		Bridging the Visual-to-Physical Gap: Physically Aligned Representations for Fall Risk Analysis
	}
	
	\author[\firstAuthorLast ]{\Authors} %This field will be automatically populated
	\address{} %This field will be automatically populated
	\correspondance{} %This field will be automatically populated
	
	\extraAuth{}% If there are more than 1 corresponding author, comment this line and uncomment the next one.
	%\extraAuth{corresponding Author2 \\ Laboratory X2, Institute X2, Department X2, Organization X2, Street X2, City X2 , State XX2 (only USA, Canada and Australia), Zip Code2, X2 Country X2, email2@uni2.edu}

	\maketitle
	\onehalfspacing

\begin{abstract}
Vision-based fall analysis has advanced rapidly, but a key bottleneck remains: visually similar motions can correspond to very different physical outcomes because small differences in contact mechanics and protective responses are hard to infer from appearance alone.
Most existing approaches handle this by supervised injury prediction, which depends on reliable injury labels. In practice, such labels are difficult to obtain: video evidence is often ambiguous (occlusion, viewpoint limits), and true injury events are rare and cannot be safely staged, leading to noisy supervision.
We address this problem with PHARL (PHysics-aware Alignment Representation Learning), which learns physically meaningful fall representations without requiring clinical outcome labels.
PHARL regularizes motion embeddings with two complementary constraints: (1) trajectory-level temporal consistency for stable representation learning, and (2) multi-class physics alignment, where simulation-derived contact outcomes shape embedding geometry.
By pairing video windows with temporally aligned simulation descriptors, PHARL captures local impact-relevant dynamics while keeping inference purely feed-forward.
Experiments on four public datasets show that PHARL consistently improves risk-aligned representation quality over visual-only baselines while maintaining strong fall-detection performance. Notably, PHARL also exhibits \emph{zero-shot ordinality}: an interpretable severity structure (Head $\succ$ Trunk $\succ$ Supported) emerges without explicit ordinal supervision.
\end{abstract}

\noindent{\keyFont Keywords:} fall risk analysis, representation learning, contrastive learning, embedding geometry

\section{Introduction}
Falls are complex events involving fast posture transitions and contact interactions. Although computer vision has made strong progress in binary fall detection \citep{gutierrez2021comprehensive, benkaci2024vision}, a central challenge remains: \emph{kinematic-to-physical ambiguity}. In practice, trajectories that look similar can still lead to very different outcomes, depending on subtle protective responses and contact topology (e.g., arm-braced landing versus direct head impact) \citep{robbins2014impact, choi2014head}. For purely data-driven models, separating these cases is difficult despite their markedly different physical consequences.

Recent developments in this field have mainly focused on stronger spatiotemporal modeling and large-scale representation learning. CNN/LSTM pipelines established competitive early baselines \citep{islam2024deep}, while graph-based and transformer-based architectures further improved detection robustness in realistic settings \citep{ma2024stgcn, chen2025unigcn, liu2024transformer}. In parallel, contrastive and masked pretraining paradigms have significantly improved motion representation quality in action understanding \citep{dave2021tclr, qian2021spatiotemporal, su2024motionbert, lin2024skeletonmae, duan2024revisiting}. However, most of these advances still optimize visual or temporal discrimination, and therefore provide limited supervision for distinguishing physically distinct outcomes within the same ``fall'' category.

Beyond binary detection, current research has started to address fall risk and post-fall consequence analysis, including gait-based risk assessment \citep{moore2024enhancing}, privacy-preserving anomaly modeling \citep{ahmed2024unsupervised}, and sensor-fusion or skeleton-centric pipelines for robustness in real environments \citep{wang2020fall, li2020fall}. Nevertheless, most practical systems are still optimized for event detection rather than physically grounded outcome characterization. As a result, controlled low-impact falls and high-impact falls may still be represented too similarly when supervision is limited to coarse labels.

Traditional outcome-oriented strategies attempt to predict injury severity directly from labeled data. Although intuitive, this route depends on reliable injury annotations, which are difficult to acquire in realistic settings. Visual evidence is often ambiguous because of occlusions and viewpoint limitations, and genuine injury outcomes are rare and cannot be safely staged. As a result, direct supervision is inherently noisy. Motivated by this limitation, we shift the objective from outcome prediction to representation shaping: the embedding should encode physics-consistent contact structure so that downstream analysis can separate outcome-relevant differences without dense clinical labels.

To this end, we introduce PHARL (PHysics-regularized Alignment Representation Learning), which uses physics simulation as structural supervision rather than as a direct prediction target. Concretely, estimated human motion is retargeted to a high-fidelity humanoid model, and short-horizon simulation produces coarse contact outcomes. These outcomes define cross-trajectory equivalence relations that regularize latent-space geometry through multi-class physics alignment. This reduces motion-outcome ambiguity and reveals an interpretable zero-shot ordinal trend.

This design is intentionally non-predictive: PHARL does not output clinical diagnosis, injury probability, or deployable severity scores. Instead, physics-derived outcomes---obtained without manual injury annotation---are only used to build contrastive relations and neighborhood constraints in representation space. Compared with purely motion-driven objectives, this formulation better separates physically distinct cases (e.g., successful upper-limb protection versus direct head-ground contact), while preserving efficient feed-forward inference because simulation is only required during training.

The main contributions of this paper are summarized as follows:
\begin{enumerate}
\item We reformulate fall analysis as a weakly supervised representation learning problem, where physics-derived contact outcomes provide structural supervision without training an explicit outcome predictor.
\item We propose a physics-consistency objective that regularizes embedding geometry through denominator masking (false-negative removal) and auxiliary physics-aligned positives, using physics-derived equivalence instead of manual outcome labels.
\item We establish a rank-prioritized evaluation protocol to assess both utility and interpretability, including zero-shot ordinality (Spearman $\rho$, POA), contact localization (Binary Contact AP/AUC), fall-detection sanity checks (AUC), and geometric diagnostics (PCR, Kendall $\tau$).
\end{enumerate}

The remainder of this paper is organized as follows. 
Section~\ref{sec:related_work} reviews related work. 
Section~\ref{sec:method} presents the PHARL framework, including motion-level temporal consistency and physics-level outcome consistency. 
Section~\ref{sec:experiments} describes the experimental setup and results. 
Section~\ref{sec:discussion} discusses implications and limitations, and 
Section~\ref{sec:conclusion} concludes the paper.

\section{Related Work}\label{sec:related_work}

\subsection{Fall Detection and Outcome Analysis}
Vision-based fall analysis has evolved from handcrafted descriptors to deep spatiotemporal modeling. 
Early studies formulated fall detection as action classification with manually designed motion features \citep{turaga2008machine, zhang2013fall, de2016optimized}. 
With the growth of public datasets such as Le2i, URFD, MCFD, and CAUCAFall \citep{Le2i, URFD, MCFD, CAUCAFall}, deep-learning based methods became the dominant paradigm for capturing scene context and temporal cues under more realistic conditions.

Recent surveys report consistent gains from CNN/LSTM pipelines and multimodal fusion strategies, especially in complex daily-activity backgrounds \citep{wang2020fall, gutierrez2021comprehensive, benkaci2024vision, islam2024deep}. 
Skeleton-centered approaches further improve robustness by reducing appearance bias and focusing on motion structure \citep{li2020fall, ma2024stgcn, chen2025unigcn}. 
Transformer-based architectures then extend this direction by modeling long-range dependencies and reducing false alarms caused by visually similar non-fall activities \citep{liu2024transformer}.

Despite these advances, most systems are still optimized for binary event recognition (fall vs. non-fall), where physically different post-fall outcomes may share the same supervision signal. 
In practical safety monitoring, this leaves a gap between detection capability and outcome-aware interpretation, particularly when only weak labels are available.

A parallel line of research targets fall risk estimation and prevention rather than post-fall representation. 
Representative efforts include gait-informed risk assessment \citep{amundsen2022fall, wang2023gait}, anomaly-based monitoring under privacy constraints \citep{ahmed2024unsupervised}, and clinical safety frameworks emphasizing incident prevention and system-level risk management \citep{WHO2021, ecri2025, london2024}. 
These studies are crucial for prevention, but they generally do not learn contact-aware latent structures for post-fall motion outcome analysis.

\subsection{Contrastive Representation Learning for Human Motion}
Self-supervised contrastive learning has become a core strategy for representation learning without dense annotation \citep{chen2020simple, he2020momentum, oord2018representation}. 
In video and motion understanding, temporal consistency and instance discrimination objectives improve transferability across downstream tasks \citep{dave2021tclr, qian2021spatiotemporal}. 
Recent methods extend these ideas with harder negative mining and fine-grained positive construction to improve intra-class discrimination \citep{kalantidis2020hard, jiang2024fine}.

For skeleton and human-motion domains, large-scale pretraining methods such as MotionBERT \citep{su2024motionbert} and SkeletonMAE \citep{lin2024skeletonmae} report strong gains in representation quality. 
Related frameworks including multi-skeleton and mask-based contrastive variants further demonstrate that objective design substantially influences learned motion semantics \citep{guo2024msclr, yang2024cml}. 
Benchmark studies and surveys confirm that the field is moving toward broader pretraining and stronger invariance learning \citep{shah2024anubis, duan2024revisiting, yan2024skeleton}.

However, the invariances encouraged by standard contrastive objectives are not always aligned with fall-outcome understanding. 
As discussed by \cite{wang2024view}, augmentations tailored to viewpoint invariance may suppress physically informative cues. 
Moreover, temporal proximity alone can create positive pairs that are visually similar but outcome-inconsistent, which limits direct applicability to contact-sensitive fall representation.

\subsection{Physics-Informed Learning for Human Dynamics}
Physics-informed learning introduces domain priors to improve plausibility and consistency in human motion modeling. 
Broadly, existing methods rely on soft constraints (e.g., penetration/contact penalties) or hard constraints through differentiable simulation and dynamics optimization \citep{karniadakis2021physics, rempe2021humor, xie2021physics}. 
In synthesis and forecasting tasks, physics-guided frameworks such as PhysDiff \citep{yuan2023physdiff} and PIMNet \citep{zhang2022pimnet} show clear improvements in realism and physical feasibility.

Recent studies also explore integrating world models, physics simulators, and embodied priors into perception pipelines \citep{shen2024world, shen2025multi, zhang2023patch}. 
These methods demonstrate that simulation signals can regularize representations beyond pure appearance cues, especially when real annotations are sparse.

Compared with prior physics-informed work, PHARL does not use physics to generate trajectories at deployment time, nor to train an end-to-end outcome classifier. 
Instead, simulation is used during training to construct weak structural supervision in embedding space. 
This design preserves inference efficiency while injecting contact-relevant physical structure.

In summary, existing fall-detection literature provides strong event-level discrimination, contrastive learning offers scalable representation learning, and physics-informed methods contribute physical plausibility. 
Yet their intersection remains underdeveloped for outcome-aware fall representation under scarce clinical labels. PHARL targets this intersection by combining contrastive learning with simulation-derived outcome equivalence, aiming to improve contact-aware organization without requiring injury labels or simulation at inference time.

\section{Method}\label{sec:method}

\subsection{Problem Formulation}
We study outcome-aware representation learning for fall videos without injury annotations. 
Let $\set{X} = \{x_1, x_2, \dots, x_N\}$ denote a dataset of RGB clips depicting falls or daily activities. 
Each clip $x_i$ has $T$ frames, but no clinical injury label $y_i^{\text{injury}}$ is available. 
3D pose/SMPL information is used only offline to run simulation and construct physics/temporal relations; it is never used as encoder input and is unavailable at inference time.

Our goal is to learn an encoder $f: x_i \mapsto \vect{z}_i \in \mathbb{R}^d$ that maps motion observations to embeddings satisfying three properties:
\begin{enumerate}
	\item Motion coherence: Temporally adjacent clips from the same sequence map to nearby embeddings, ensuring temporal continuity.
	\item Outcome consistency: Clips leading to similar physics-derived contact outcomes map to nearby embeddings, even if their visual trajectories differ.
	\item Non-predictive semantics: The representation encodes outcome-aware structure without training an outcome predictor or enforcing ordinal ranking; any ordering observed in the embedding space is evaluated post-hoc for analysis only.\footnote{We use ``outcome'' to refer to physics-derived contact configurations (\textit{Supported, Trunk, Head}), not clinical injury severity or severity scores.}
\end{enumerate}

PHARL is formulated at the temporal-window level rather than the full-video level. 
We adopt this setting because a single fall trajectory can contain multiple physical phases (pre-impact, trunk impact, head impact), which are difficult to represent with one video-level label.

These properties must emerge without direct supervision on injury outcomes. 
In practice, training uses only trajectory identity and physics-derived weak labels; no clinical target is optimized. 
Consequently, evaluation can only assess representation structure---such as contact-aware separability and physics-consistent geometry---rather than predictive performance on injury classification.

\begin{figure}[t]
	\begin{center}
		\includegraphics[width=.98\linewidth]{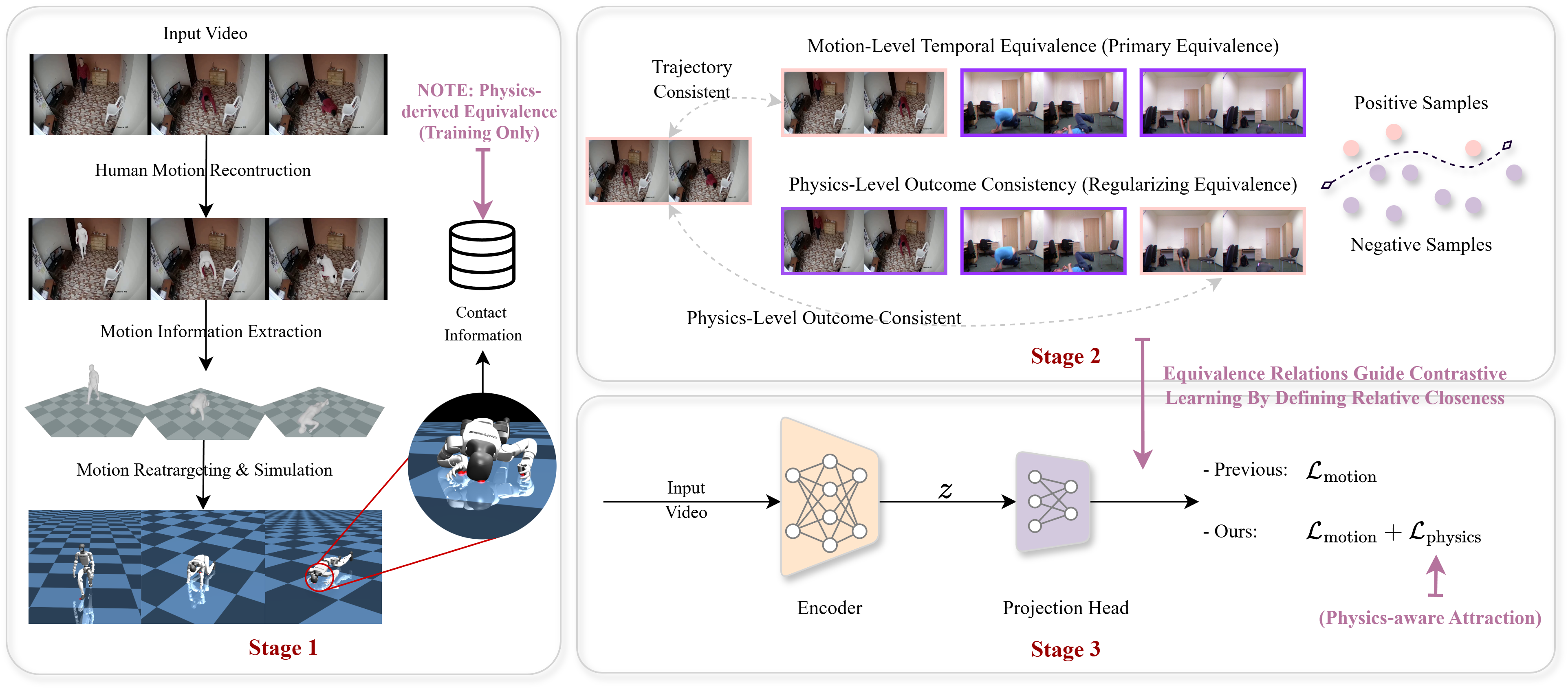}
	\end{center}
	\caption{
		Overview of PHARL. 
		Stage 1 (training only): 
			RGB videos are processed offline to reconstruct motion and run physics simulation, producing window-level contact outcomes (\textit{Supported, Trunk, Head}). 
		Stage 2: 
			PHARL applies two complementary constraints: trajectory-consistent temporal positives and physics-level contact structure across trajectories. Physics structure is used both for denominator masking in the trajectory loss and for auxiliary same-class attraction among contact windows. 
		Stage 3: 
			An encoder is trained with a composite objective (Eq.~(\ref{eq_3})) that improves contact-aware geometry without adding an outcome prediction head. During inference, PHARL uses only the feed-forward RGB encoder, with no physics simulation.
	}
	\label{fig:pipeline}
\end{figure}

\subsection{Overview of PHARL Framework}
The PHARL framework contains three components (Figure~\ref{fig:pipeline}): 
(1) a motion representation encoder, instantiated as a physics-free baseline and a physics-regularized variant under the same architecture; 
(2) a physics module that converts reconstructed motion into window-level contact outcomes and relation metadata;
(3) learning objectives that jointly enforce temporal coherence and physics-level outcome consistency.

Both encoder variants share the same backbone, so performance differences arise from supervision design rather than model capacity. 
The baseline is trained with motion-level consistency only, whereas the PHARL variant adds physics-guided regularization. 
Physics information is required only during training; at inference, both models run as standard feed-forward RGB encoders without simulation overhead.

\textbf{Boundary clarification.} 
(1) Input: PHARL takes raw RGB clips as model input. 
(2) Training-only metadata: 3D human pose/SMPL estimates are extracted offline for simulation and relation construction, and are never fed into the encoder. 
(3) Supervision type: Physics labels are weak structural signals (for geometric constraints), not pseudo-label targets for an outcome prediction head. 
(4) Terminology: We distinguish between contacts observed within the current window and contacts observed from a short-horizon continuation initialized at the window boundary. Both are mapped to the same coarse outcome categories \(y^{\text{phys}}\in\{\textit{Supported, Trunk, Head}\}\). Throughout the manuscript, ``outcome'' refers to these physics-defined contact categories (not clinical injury severity). The short-horizon continuation is included to capture immediate post-boundary impacts that may not yet appear in the window itself.

\subsection{Motion-Level Temporal Consistency}\label{sec:motion_equiv}
We use a trajectory-based contrastive objective to preserve motion coherence. 
For anchor embedding $\vect{z}_i$, positives are other views/windows from the same trajectory, while samples from other trajectories serve as contrastive negatives. The motion term is
\begin{equation}\label{eq_1}
	\mathcal{L}_{\text{motion}} =
		-\log
		\frac{\sum_{j \in \mathcal{P}_i^{\text{traj}}}\exp\!\big(\simop(\vect{z}_i,\vect{z}_j)/\tau\big)}
		{\sum_{k \in \mathcal{A}_i}\exp\!\big(\simop(\vect{z}_i,\vect{z}_k)/\tau\big)},
\end{equation}
where $\vect{z}_i \in \mathbb{R}^d$ is the $\ell_2$-normalized embedding of anchor $i$, $\simop(\cdot,\cdot)$ denotes cosine similarity, $\tau$ is the motion-branch temperature, $\mathcal{P}_i^{\text{traj}}$ is the trajectory-positive set for anchor $i$, and $\mathcal{A}_i$ is the candidate set excluding self-comparisons.

This term stabilizes temporal representation learning, but it is insufficient for contact-state awareness: clips from different trajectories may appear visually similar while leading to different contact consequences. 
PHARL addresses this limitation by injecting physics-derived structure into training.
In essence, the motion objective provides a temporal prior, and the physics objective provides structural correction. 
PHARL therefore refines motion representation rather than replacing it with outcome prediction.

\subsection{Physics-Level Outcome Consistency}

\subsubsection{Outcome Structure as Weak Supervision}
This module injects physics-consistent structure into the embedding space without introducing an outcome classifier. 
Each window receives a coarse physics label \(y^{\text{phys}} \in \{\textit{Supported, Trunk, Head}\}\), which is used only to define geometric constraints in contrastive learning.
PHARL uses this supervision in two complementary ways:
(1) binary contact grouping (\textit{Head/Trunk} vs. \textit{Supported}) for denominator masking, and
(2) exact class matching (\textit{Trunk-to-Trunk, Head-to-Head}) for auxiliary attraction.
The first reduces harmful repulsion among contact windows; the second strengthens cross-trajectory alignment for contact-consistent events. 
Together, they increase outcome-aware organization under weak supervision while preserving feed-forward RGB inference.

\subsubsection{Outcome Extraction and Denoising}
Physics-derived descriptors are used as weak structural supervision rather than prediction targets. 
The purpose of denoising is to resolve the \textit{visual-semantic gap}—a fundamental source of noise where temporally mismatched labels contaminate the representation. 
For instance, in traditional trajectory-level labeling, every window of a fall video inherits the same label, forcing the encoder to associate pre-impact standing poses with severe outcomes (e.g., ``Head Impact''). 
PHARL's denoising process produces labels that are temporally aligned with each window and robust to transient simulation noise.

For each video window \(\mathcal{W}=[t_0,t_1)\), label construction follows three steps.
First, temporal alignment retains only contact descriptors whose interval \([t_s,t_e)\) overlaps \(\mathcal{W}\), i.e., \((t_s<t_1)\land(t_e>t_0)\). 
This step correctly re-assigns pre-impact frames to the \textit{Supported} class, ensuring that severe labels are reserved for actual impact kinematics.
Second, boundary completion incorporates short-horizon continuation evidence from the window endpoint, so that impacts that occur immediately after \(t_1\) can still inform the current state.
This step addresses boundary truncation in fixed-length windows, where pre-impact frames and impact onset may be split across adjacent windows.
Third, reliability filtering suppresses weak impulses and aggregates the remaining evidence by taking the category-wise maximum impulse across available sources.
These operations define a deterministic, reproducible mapping from raw contact descriptors to window-level supervision.

Let the filtered impulse set for \(\mathcal{W}\) be \(\mathcal{S}_{\mathcal{W}}\). 
The coarse physics label is assigned by hierarchical dominance:
This priority rule is physics-motivated: head contact is treated as the most critical outcome state, followed by trunk contact, while all remaining cases are mapped to \textit{Supported}.
\begin{equation}\label{eq_2}
	y^{\text{phys}}(\mathcal{W})=
		\begin{cases}
		\text{\textit{Head}}, & \text{if head-contact impulse is present in } \mathcal{S}_{\mathcal{W}},\\
		\textit{Trunk}, & \text{if torso/hip-contact impulse is present in } \mathcal{S}_{\mathcal{W}},\\
		\textit{Supported}, & \text{otherwise (including arm/hand/leg/foot contacts).}
		\end{cases}
\end{equation}
If no valid overlapped descriptor is available (or the window is non-fall), the label defaults to \textit{Supported}.

In this context, denoising provides stable weak supervision for physics regularization without introducing an explicit outcome-prediction head. It ensures the representation geometry is shaped by temporally localized physical evidence rather than coarse video-level proxies.

\subsubsection{Representation Learning Objectives}
The physics-regularized encoder combines trajectory consistency with outcome-aware regularization:
\begin{equation}\label{eq_3}
    \mathcal{L} =
	    \mathcal{L}_{\text{motion}}
	    + \lambda_{\text{phys}}\mathcal{L}_{\text{physics}}
	    + \lambda_{\text{var}}\mathcal{L}_{\text{var}}.
\end{equation}
Here, $\mathcal{L}_{\text{motion}}$ is the trajectory-consistency term, $\mathcal{L}_{\text{physics}}$ is the physics-alignment term, and $\mathcal{L}_{\text{var}}$ is the variance regularizer that prevents representation collapse. $\lambda_{\text{phys}}$ and $\lambda_{\text{var}}$ are balancing coefficients.

For denominator masking, we define a binary contact indicator
$C_i=\mathbbm{1}[y_i^{\text{phys}}\in\{\textit{Head, Trunk}\}]$ and the masked set
\begin{equation}\label{eq_4}
    \mathcal{M}_i=\{k\in \mathcal{A}_i \mid C_i=1,\; C_k=1,\; \text{traj}(k)\neq \text{traj}(i)\},
\end{equation}
where $C_i$ is the binary contact indicator for anchor $i$ and $\mathcal{M}_i$ is the masked candidate set that removes cross-trajectory contact windows from the denominator. If $C_i=0$ (\textit{Supported} anchor), then $\mathcal{M}_i=\varnothing$.
The masked trajectory loss is
\begin{equation}\label{eq_5}
    \mathcal{L}_{\text{motion}} =
	    -\log
	    \frac{\sum_{j\in \mathcal{P}_i^{\text{traj}}}\exp(\simop(\vect{z}_i,\vect{z}_j)/\tau)}
	    {\sum_{k\in \mathcal{A}_i \setminus \mathcal{M}_i}\exp(\simop(\vect{z}_i,\vect{z}_k)/\tau)},
\end{equation}
where denominator candidates are restricted to $\mathcal{A}_i \setminus \mathcal{M}_i$, i.e., all available comparisons after masking.
Without this masking step, contact windows from different trajectories can be treated as hard negatives, which introduces false repulsion and weakens contact-consistent geometry.

For auxiliary attraction, contact anchors use exact class matching across trajectories:
\begin{equation}\label{eq_6}
    P_i^{phys}=\{j \mid y_j^{\text{phys}}=y_i^{\text{phys}},\; y_i^{\text{phys}}\in\{\textit{Head, Trunk}\},\; \text{traj}(j)\neq\text{traj}(i)\}.
\end{equation}
Here, $P_i^{phys}$ denotes same-class (\textit{Head, Trunk}) cross-trajectory positives for anchor $i$, and $Q_i=\{k \mid \text{traj}(k)\neq\text{traj}(i)\}$ denotes all cross-trajectory candidates for anchor $i$. The auxiliary term is
\begin{equation}\label{eq_7}
    \mathcal{L}_{\text{physics}} =
	    -\log
	    \frac{\sum_{j\in P_i^{phys}}\exp(\simop(\vect{z}_i,\vect{z}_j)/\tau_p)}
	    {\sum_{k \in Q_i}\exp(\simop(\vect{z}_i,\vect{z}_k)/\tau_p)},
\end{equation}
where $\tau_p$ is the physics-branch temperature.
Without this exact-class attraction term, the embedding tends to preserve coarse contact/non-contact separation but under-represents fine-grained ordinal structure between \textit{Trunk} and \textit{Head}.

While the scarcity of severe events (\textit{Head}) often leads to their merging with \textit{Trunk} into a single binary category, we explicitly enforce exact multi-class attraction. This design choice is important for revealing the zero-shot ordinality of the manifold; by pulling only exact matches together, the model is encouraged to learn distinct kinematic signatures associated with different physical energy levels.
In practice, anchors with empty positive sets (e.g., no available cross-trajectory same-class partner) are skipped for the corresponding term, following standard contrastive training conventions.

This formulation encourages contact-consistent geometry without introducing an outcome prediction head or ordinal supervision.
Each term serves a specific role: 
$\mathcal{L}_{\text{motion}}$ (with masking) preserves trajectory-level temporal consistency while reducing unnecessary repulsion among contact windows; 
$\mathcal{L}_{\text{physics}}$ strengthens cross-trajectory alignment for matched outcome classes (\textit{Head, Trunk}); 
and 
$\mathcal{L}_{\text{var}}$ maintains embedding spread to prevent collapse.

\subsection{Training and Inference Workflow}
For clarity, we summarize PHARL training as a four-step pipeline (input/output view). 
(1) Window construction: each trajectory is segmented into overlapping windows (input: RGB trajectory; output: window set). 
(2) Physics signal extraction (training only): each window is associated with simulation-derived contact outcomes and contact statistics, with short-horizon continuation used to recover immediate post-boundary contacts (output: window-level physics metadata). 
(3) Relation building: trajectory-positive sets, denominator masks, and cross-trajectory same-class sets are built from window metadata (output: contrastive relation graph). 
(4) Joint optimization: the encoder is updated with Eq.~(\ref{eq_3}), combining temporal consistency and physics-alignment regularization (output: trained RGB encoder).

Algorithmically, the key distinction from standard temporal contrastive learning is that negative/positive relations are no longer determined only by trajectory identity or augmentation pairing. 
PHARL uses physics equivalence to suppress harmful negatives (contact-contact false repulsion) and to add structured positives (same-class cross-trajectory attraction), thereby improving outcome-aware geometry under weak supervision.

At inference, the pipeline is simplified to a single feed-forward RGB encoder; simulation, contact extraction, and relation construction are all removed. 
Therefore, PHARL preserves deployment efficiency while retaining the physically informed structure learned during training.

\section{Experiments}\label{sec:experiments}

\subsection{Datasets}
We evaluate PHARL on four public fall datasets and aggregate them into a unified benchmark spanning multiple scenes, viewpoints, and subjects. 
Table~\ref{tab:datasets} reports trajectory-level statistics.

\begin{table}[t]
	\centering
	\caption{Dataset Information.}
	\label{tab:datasets}
	\renewcommand{\arraystretch}{1.3}
	\small
	\begin{tabular}{lrrr}
		\toprule
		Dataset & Total & Fall & No-fall \\
		\midrule
		CAUCAFall \citep{CAUCAFall}  & 100 & 50  & 50 \\
		GMDCSA-24 \citep{GMDCSA-24}  & 160 & 79  & 81 \\
		Le2i \citep{Le2i}            & 190 & 130 & 60 \\
		URFD \citep{URFD}            & 100 & 60  & 40 \\
		\midrule
		Total & 550 & 319 & 231 \\
		\bottomrule
	\end{tabular}
\end{table}

Le2i \citep{Le2i} contains 190 indoor sequences. 
URFD \citep{URFD} provides 100 sequences of falls and daily activities from a single-view setup.
CAUCAFall \citep{CAUCAFall} includes 100 home-environment sequences captured from multiple camera angles. 
GMDCSA-24 \citep{GMDCSA-24} contributes 160 trajectories with diverse fall directions and protective responses. 

None of the datasets provides clinical injury annotations. 
Binary fall labels are used only for aggregation and baseline diagnostics, not as supervision for physics-regularized representation learning. 
We perform trajectory-level train/val/test splitting to avoid window leakage, with stratified sampling to preserve head-contact trajectories across splits. 
The final split is 438/56/56 trajectories (train/val/test). 
Physics contact descriptors are precomputed and cached per video, and head-contact flags from these descriptors are used only for split stratification. 
This protocol enforces strict trajectory-level separation throughout the pipeline.

Each video is converted into overlapping windows using a sliding-window strategy. 
For each window, we derive a coarse physics outcome label $y^{\text{phys}}\in\{\textit{Supported},\textit{Trunk},\textit{Head}\}$ from contact statistics.
These labels are used only for training-time regularization and post-hoc evaluation, never as encoder input or prediction targets. 
Window labels are computed with a minimum impulse threshold of 0.0~N$\cdot$s; video and physics windows are aligned by temporal overlap. 
Depending on the ablation setting, labels are derived from in-window contacts, short-horizon continuation contacts, or their union; when both are used, sources are merged by per-category maximum impulse. The training split is imbalanced (\textit{Supported} $\approx 56\%$, \textit{Trunk} $\approx 34\%$, \textit{Head} $\approx 10\%$), with similar class trends on validation and test sets.

\subsection{Evaluation Metrics}\label{sec:metrics}
We design an evaluation suite to test whether the learned embedding geometry is consistent with physics-derived outcomes, without using clinical injury labels.
All reported metrics are post-hoc diagnostics.
Model checkpoints are selected solely based on the minimum validation loss under the training objective.
Throughout this section, ``outcome'' refers to physics-derived contact categories (\textit{Supported, Trunk, Head}). 
These categories are analysis labels, not optimization targets.

We define a post-hoc severity axis after training using train-split centroids:
\begin{equation}\label{eq_8}
	\begin{aligned}
		\hat{\mathbf{v}} 
		&= \frac{\boldsymbol{\mu}_{\textit{Head}} - \boldsymbol{\mu}_{\text{\textit{Supported}}}}
		{\left\lVert \boldsymbol{\mu}_{\textit{Head}} - \boldsymbol{\mu}_{\text{\textit{Supported}}} \right\rVert}
		\quad \text{(severity axis)}, \\
		s_i 
		&= \mathbf{z}_i^\top \hat{\mathbf{v}}
		\quad \text{(projection score)} .
	\end{aligned}
\end{equation}
Here, $\boldsymbol{\mu}_{\textit{Head}}$ and $\boldsymbol{\mu}_{\text{\textit{Supported}}}$ denote class centroids (computed on the training set), and $s_i$ denotes the scalar projection of embedding $\mathbf{z}_i$ onto the post-hoc severity axis. This projection score is used only for post-hoc analysis and is not a prediction output.
Using ordinal labels $\{0,1,2\}$ for \{\textit{Supported, Trunk, Head}\}, we report metrics in the following priority order:
\begin{enumerate}
	\item Spearman's $\rho$: rank correlation between ordinal labels and projection scores $s_i$, computed on their respective ranks, reflecting global monotonic alignment along the post-hoc severity axis.
	
	\item POA (Macro): macro-averaged pairwise ordering accuracy over sampled instance pairs, measuring whether higher-severity outcomes are consistently ranked above lower-severity ones.
	
	\item Binary Contact AP: average precision of a linear probe distinguishing contact (\textit{Head, Trunk}) from \textit{Supported} cases, emphasized under class imbalance.
	
	\item Binary Contact AUC: ROC-AUC of the same linear probe, capturing threshold-free separability between contact and \textit{Supported} instances.
	
	\item Fall Detection AUC: ROC-AUC of a linear probe for fall versus no-fall, reported solely as a deployment-oriented sanity check.
	
	\item Physics Consistency Ratio (PCR): ratio of inter-class to intra-class embedding distances defined by $y^{\text{phys}}$, indicating geometry-level separation consistent with physics-derived categories.
	
	\item Kendall's $\tau$: concordance-based rank correlation between ordinal labels and projection scores, reported as a robustness check for ordinal trends.
\end{enumerate}

For interpretation, we prioritize metrics as follows: 
Spearman’s $\rho$ and POA provide primary evidence of ordinal consistency; 
Binary Contact AP/AUC assess contact-localization utility; 
and Fall Detection AUC, PCR, and Kendall’s $\tau$ are secondary diagnostics for deployment sanity and geometry quality.
%No evaluation metric is used for checkpoint selection.

\begin{table}[t]
	\centering
	\caption{
		Performance comparison. Bold indicates the best result and underlined values indicate the second-best result.
	}
	\label{tab:baseline_results}
	\renewcommand{\arraystretch}{1.0}
	\small
	\begin{tabular}{l @{\hspace{6pt}} ccccccc}
		\toprule
		\multirow{2}{*}{Method} & \multirow{2}{*}{Spearman \(\rho\)} & \multirow{2}{*}{POA (Macro)} & \multicolumn{2}{c}{Binary Contact} & \multirow{2}{*}{Fall Detection AUC} & \multirow{2}{*}{PCR} & \multirow{2}{*}{Kendall \(\tau\)} \\
		\cmidrule(lr){4-5}
		&  &  & AP & AUC &  &  &  \\
		\midrule
		Vanilla        & 0.2232 & 0.6221 & 0.4992 & 0.7056 & 0.7736 & 1.0082 & 0.1768 \\
		HNM            & 0.2454 & 0.6234 & 0.5979 & 0.7762 & 0.8081 & 1.0032 & 0.1951 \\
		BT             & 0.2405 & 0.6190 & 0.5979 & 0.7688 & 0.8289 & 1.0055 & 0.1911 \\
		SimSiam        & 0.0396 & 0.5199 & 0.3792 & 0.5819 & 0.5667 & \textbf{1.0390} & 0.0315 \\
		CD             & 0.2189 & 0.6255 & 0.5429 & 0.7476 & 0.8302 & 1.0076 & 0.1736 \\
		MGC            & \underline{0.2754} & \underline{0.6382} & \underline{0.6452} & \underline{0.7987} & \underline{0.8405} & 1.0046 & \underline{0.2189} \\
		\midrule
		PHARL     & \textbf{0.4800} & \textbf{0.7983} & \textbf{0.6484} & \textbf{0.8223} & \textbf{0.8996} & \underline{1.0126} & \textbf{0.3850} \\
		\bottomrule
	\end{tabular}
\end{table}

\subsection{Baselines}
We compare PHARL against six baselines using identical data splits.
All methods rely on raw RGB clips with trajectory-consistent temporal positives for stable optimization.
Only PHARL incorporates physics-derived structural supervision.
This deliberate information asymmetry isolates the effect of physics regularization, rather than asserting superiority under equivalent supervision.

\begin{itemize}
	\item Vanilla InfoNCE \citep{oord2018representation}: A standard contrastive learning baseline optimized with the InfoNCE objective, without auxiliary losses or structural constraints.
	
	\item Hard Negative Mining (HNM) \citep{kalantidis2020hard}: A contrastive variant that re-weights or emphasizes hard negative samples (i.e., negatives most similar to the anchor) to encourage finer-grained separation.
	
	\item Barlow Twins (BT) \citep{zbontar2021barlow}: A redundancy-reduction method that enforces decorrelation between embedding dimensions via a cross-correlation objective.
	
	\item SimSiam \citep{chen2021exploring}: A negative-free baseline based on asymmetric prediction and stop-gradient. In our implementation, each training item uses two augmented views of the anchor window and one additional motion-equivalent view sampled from a coarse temporal-phase bucket (early/mid/late).
	
	\item Contrastive Disentangling (CD) \citep{jiang2024fine}: A multi-head contrastive framework designed to disentangle latent factors, with auxiliary constraints promoting feature diversity.
	
	\item Multi-Grained Contrast (MGC) \citep{shen2025multi}: A temporal contrastive approach that aligns local window embeddings with a global trajectory representation to capture multi-scale temporal structure.
\end{itemize}

\subsection{Implementation Details}

We use a ResNet-18 encoder with temporal average pooling and a projection head to produce $\ell_2$-normalized embeddings. 
Physics simulation is run offline only to derive contact descriptors; physics labels are never provided as encoder inputs.

All models are trained for 100 epochs. 
We select the primary checkpoint by minimum validation loss and additionally report final-epoch checkpoints. 
Baseline and PHARL runs share learning rate $1\times10^{-4}$ and trajectory-contrast temperature $\tau=0.2$. 
PHARL further uses an auxiliary physics temperature of 0.2 with warmup-based physics-weight scheduling. 
A shared memory bank is used to stabilize contrastive optimization, and physics-stratified sampling is enabled only for PHARL variants that use physics labels. In summary, architecture, data splits, learning rate, and base temperature are shared across methods, while physics-specific settings are applied only to PHARL.

\begin{table}[t]
	\centering
	\caption{
		Ablation study of PHARL components. Bold indicates the best result and underlined values indicate the second-best result.
	}
	\label{tab:ablation_results}
	\renewcommand{\arraystretch}{1.15}
	\setlength{\tabcolsep}{3.5pt}
	\resizebox{\linewidth}{!}{
		\begin{tabular}{cccc @{\hspace{6pt}} ccccccc}
			\toprule
			\multirow{2}{*}{Denoising} & \multirow{2}{*}{Multi-Class} & \multicolumn{2}{c}{Physics Information} & \multirow{2}{*}{Spearman \(\rho\)} & \multirow{2}{*}{POA (Macro)} & \multicolumn{2}{c}{Binary Contact} & \multirow{2}{*}{Fall Detection AUC} & \multirow{2}{*}{PCR} & \multirow{2}{*}{Kendall \(\tau\)} \\
			\cmidrule(lr){3-4}\cmidrule(lr){7-8}
			&  & Window & Continuation &  &  & AP & AUC &  &  &  \\
			\midrule
			\(\times\)     & \(\checkmark\) & \(\checkmark\) & \(\checkmark\) & 0.3716 & 0.7199 & 0.5516 & 0.7747 & 0.8701 & 0.9873 & 0.2974 \\
			\(\checkmark\) & \(\times\)     & \(\checkmark\) & \(\checkmark\) & 0.4282 & 0.7685 & \underline{0.6523} & \underline{0.8249} & 0.8745 & 1.0070 & 0.3413 \\
			\(\checkmark\) & \(\checkmark\) & \(\times\)     & \(\checkmark\) & \underline{0.4708} & \underline{0.7850} & 0.6085 & 0.8146 & \textbf{0.9088} & \textbf{1.0180} & 0.3774 \\
			\(\checkmark\) & \(\checkmark\) & \(\checkmark\) & \(\times\)     & 0.3756 & 0.7272 & \textbf{0.7051} & \textbf{0.8339} & 0.8960 & \underline{1.0179} & 0.2990 \\
			\(\checkmark\) & \(\checkmark\) & \(\checkmark\) & \(\checkmark\) & \textbf{0.4800} & \textbf{0.7983} & 0.6484 & 0.8223 & \underline{0.8996} & 1.0126 & \textbf{0.3850} \\
			\bottomrule
		\end{tabular}
	}
\end{table}

\subsection{Performance Comparison}\label{sec:comparison}

Table~\ref{tab:baseline_results} reports a quantitative comparison between PHARL and representative self-supervised baselines. 
The key finding is the emergence of a clear ordinal structure in the learned representation space. 
PHARL achieves a Spearman $\rho$ of 0.4800 and a POA of 0.7983, substantially outperforming all motion-only alternatives and indicating strong alignment between latent geometry and the physics-derived risk ordering.

Standard contrastive objectives, including Vanilla InfoNCE, HNM, and BT, show limited ability to capture such ordinal relationships. 
Although these methods improve local feature discrimination and moderately increase Binary Contact AUC (e.g., 0.7762 for HNM and 0.7688 for BT), their objectives remain primarily instance-driven and provide no structural mechanism for organizing motion patterns by physical outcomes. 
Consequently, the learned embeddings exhibit weak ordinal organization, with Spearman $\rho$ values below 0.25.

Among the motion-driven baselines, MGC is the strongest. 
MGC improves ordinal correlation to $\rho=0.2754$ and achieves a Fall Detection AUC of 0.8405. 
PHARL further increases ordinal correlation to 0.4800, representing a substantial gain in ordinal alignment. 
This comparison indicates that richer motion context alone is insufficient to resolve the ambiguity of fall dynamics, whereas physics-outcome structure provides stronger guidance for representation organization.

CD adopts a multi-head representation-learning strategy but shows limited benefits in this highly imbalanced fall scenario. 
Its ordinal metrics remain comparable to those of Vanilla InfoNCE, suggesting that disentangling visual factors without physical grounding does not naturally isolate features associated with impact severity.

Finally, SimSiam exhibits clear limitations in this safety-critical setting, yielding a near-zero Spearman $\rho$ of 0.0396. 
Although it attains a relatively high PCR, the absence of negative pairs weakens the discriminative pressure needed to resolve fine-grained severity differences, resulting in representations that lack meaningful ordinal structure.

Overall, these results demonstrate that PHARL effectively mitigates the kinematic ambiguity inherent in fall motions. 
By introducing physics-outcome consistency as a structural regularizer during representation learning, PHARL consistently improves both ordinal risk alignment (Spearman, Kendall, and POA) and safety-localization performance over purely motion-driven self-supervised objectives.

\begin{figure}[t]
	\begin{center}
		\includegraphics[width=.98\linewidth]{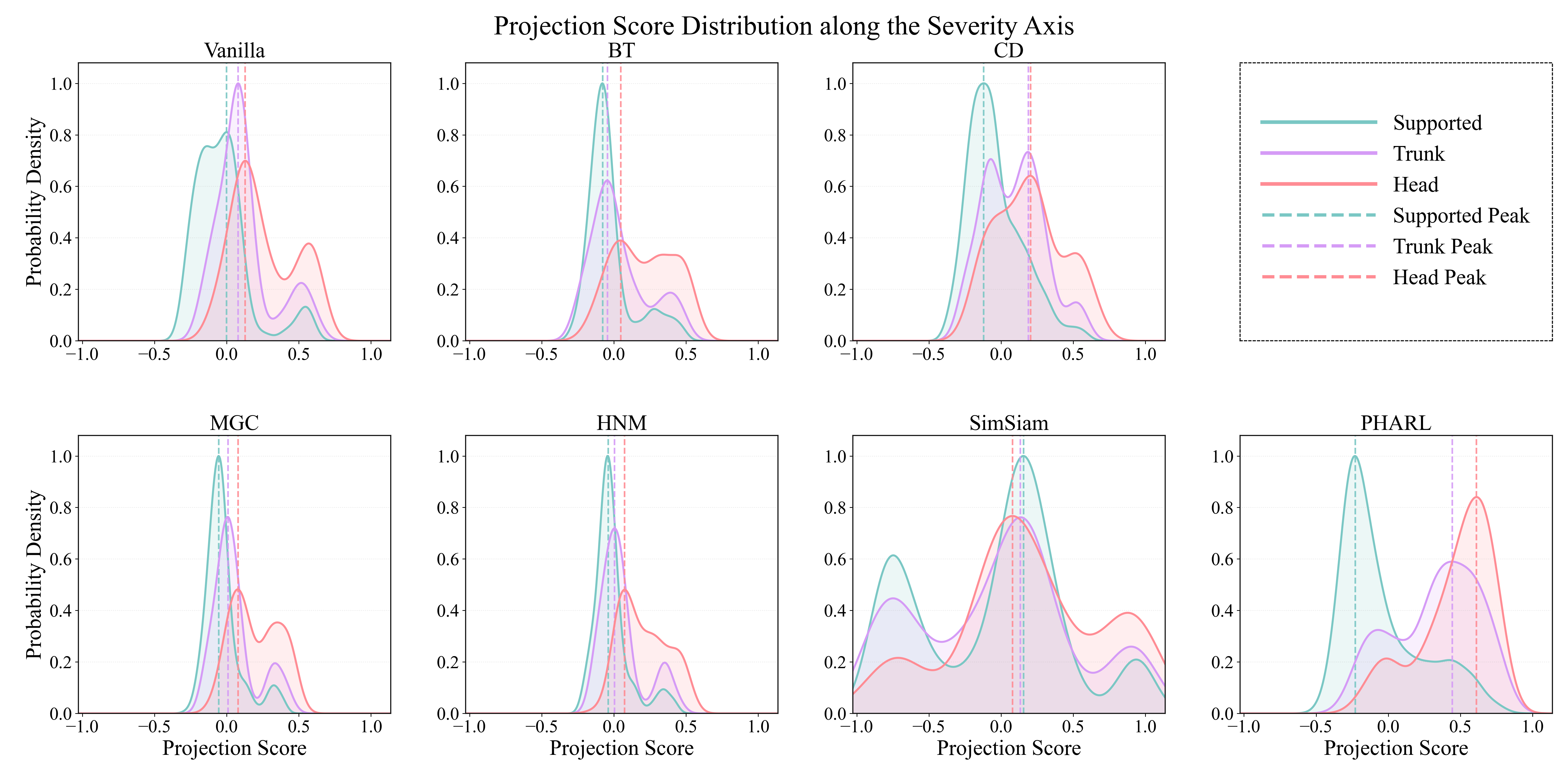}
	\end{center}
	\caption{
		Kernel density estimates of embedding distributions along the post-hoc severity axis (defined by the \textit{Supported} and \textit{Head} class centroids). Baseline methods show substantial class overlap, whereas PHARL exhibits a clearer staircase-like separation. This pattern indicates improved alignment between latent geometry and physics-consistent impact outcomes. The projection is used for analysis only and is not a prediction output.
	}
	\label{fig:manifold_kde}
\end{figure}

\subsection{Ablation Studies}\label{sec:ablation}
We conduct ablation studies to evaluate the contribution of three key components of PHARL: the denoising mechanism for window-level alignment, the multi-class attraction module for ordinally structured contrastive learning, and the two sources of physics-derived supervision (window-level contact and outcome-based continuation). 
Results are summarized in Table~\ref{tab:ablation_results}.

Removing the denoising mechanism leads to a substantial degradation in ordinal consistency. 
When trajectory-level supervision is used directly (first row), Spearman $\rho$ decreases from 0.4800 to 0.3716 and Binary Contact AP drops by nearly 10\% (0.6484 to 0.5516). 
This degradation indicates that coarse trajectory labels introduce significant label noise, where pre-impact upright frames become associated with high-severity outcomes. 
The proposed denoising step mitigates this mismatch by aligning supervision with temporally localized motion segments.

The multi-class attraction mechanism also plays an important role in shaping the severity ordering of the learned representations. 
Replacing the exact class-based attraction with binary grouping (\textit{Head} and \textit{Trunk} treated as the same class) reduces Spearman $\rho$ from 0.4800 to 0.4282 and Kendall's $\tau$ from 0.3850 to 0.3413. 
While binary supervision remains sufficient to distinguish impact events from benign activities, it fails to preserve the relative ordering among different impact severities, highlighting the importance of class-specific contrastive attraction.

The subset ablations further reveal a trade-off between instantaneous impact detection and global outcome alignment. 
Training with window-level supervision alone yields the strongest Binary Contact performance, achieving the highest AUC (0.8339) and AP (0.7051), as the supervision signal directly corresponds to observable impact frames. 
However, this localized training signal results in weaker ordinal structure, with Spearman $\rho$ dropping to 0.3756. 
In contrast, continuation-only supervision improves the global ordering of trajectories (Spearman $\rho$ of 0.4708) but slightly reduces contact localization performance.

Combining denoising, multi-class attraction, and both sources of supervision yields the most balanced overall results. 
The full PHARL configuration achieves the highest ordinal consistency (Spearman $\rho$ of 0.4800, POA of 0.7983, and Kendall's $\tau$ of 0.3850) while maintaining competitive fall detection performance (AUC of 0.8996). 
These results suggest that motion-level temporal alignment and physics-outcome consistency provide complementary supervision signals for learning risk-aware representations.

\begin{figure}[t]
	\begin{center}
		\includegraphics[width=.65\linewidth]{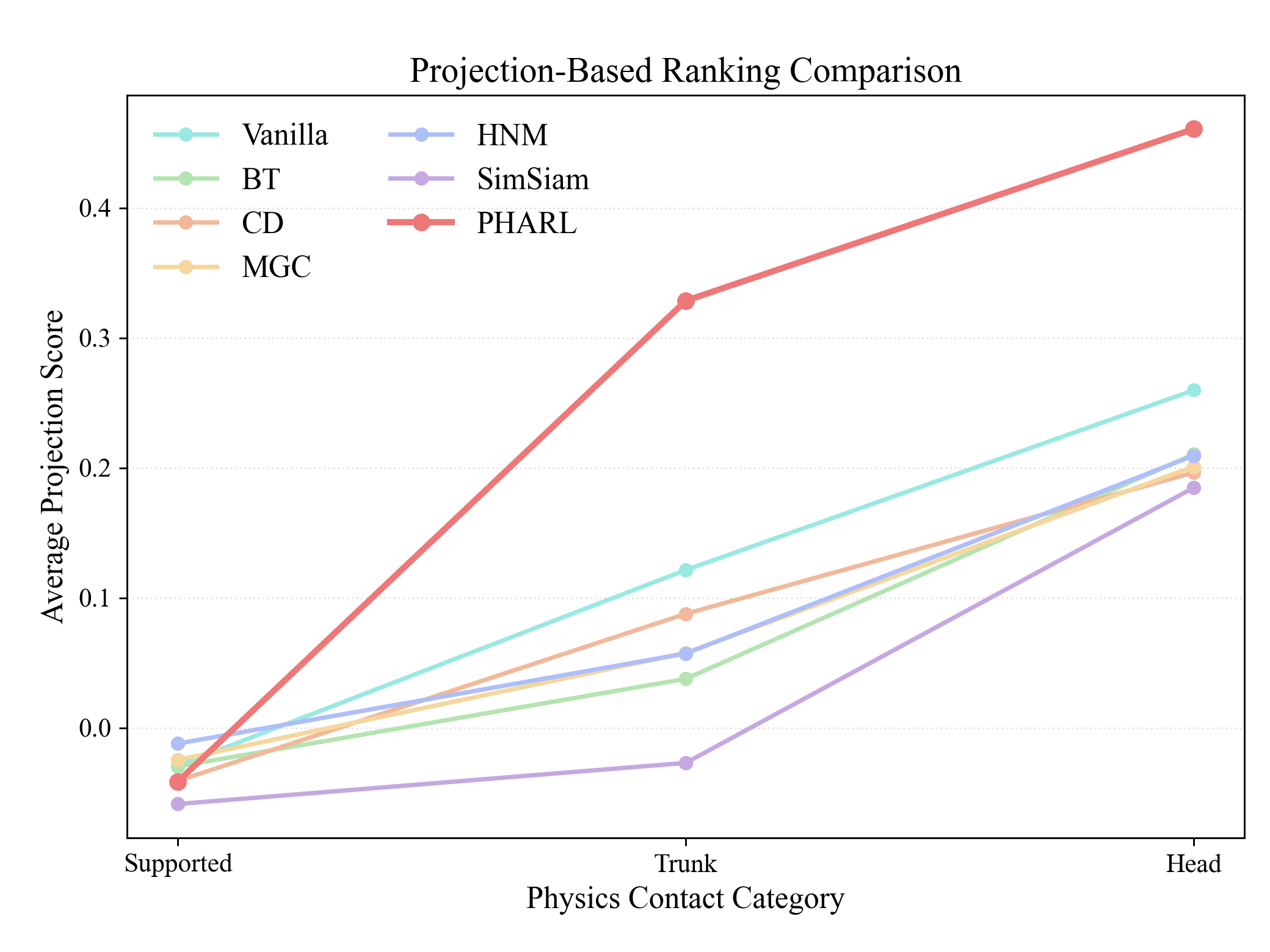}
	\end{center}
	\caption{
		Category-wise mean projection scores along the post-hoc severity axis across contact outcomes. PHARL preserves the expected physical ordering (\textit{Head $\succ$ Trunk $\succ$ Supported}) with a larger margin than motion-only baselines, indicating stronger ordinal organization in the learned representation. These projection scores are descriptive diagnostics only, not predicted severity outputs.
	}
	\label{fig:severity_summary}
\end{figure}

\subsection{Representation Analysis}\label{sec:representation_analysis}

To further examine the structural properties of the learned representation, we conduct three complementary manifold analyses focused on distribution separation, ordinal consistency, and cross-video neighborhood structure. 
These analyses provide qualitative evidence of how different methods organize impact-related motions in latent space. 
In particular, they test whether physics-regularized training induces a geometry that is consistent with the underlying ordering of physical impact outcomes.

Figure~\ref{fig:manifold_kde} visualizes the density of embeddings projected onto a post-hoc severity axis. 
This axis is defined as the unit vector from the \textit{Supported} centroid to the \textit{Head} centroid in latent space, yielding a physically interpretable direction that roughly corresponds to increasing impact severity. 
Importantly, this axis is used only for post-hoc visualization and does not affect training or quantitative metrics. 
Whereas baseline methods show substantial overlap among category distributions, PHARL produces a more structured manifold with staircase-like ordinal separation along this direction. 
Specifically, the \textit{Head} and \textit{Trunk} distributions shift rightward relative to \textit{Supported}, suggesting that the learned representation resolves kinematic ambiguity and organizes motions according to physics-consistent impact outcomes.

The emergence of ordinal structure is further examined in Figure~\ref{fig:severity_summary}, which reports the mean projection score of each category along the same severity axis. 
PHARL consistently preserves the expected physical hierarchy (\textit{Head $\succ$ Trunk $\succ$ Supported}) with a larger margin than motion-only baselines. 
This monotonic pattern indicates that physics-regularized alignment helps the encoder recover the ordinal structure of fall severity as an emergent property, despite the absence of explicit ranking supervision during training. 
Because these statistics are computed at the category level, the observed ordering is relatively insensitive to sample-count differences across categories.

\begin{figure}[t]
	\begin{center}
		\includegraphics[width=.90\linewidth]{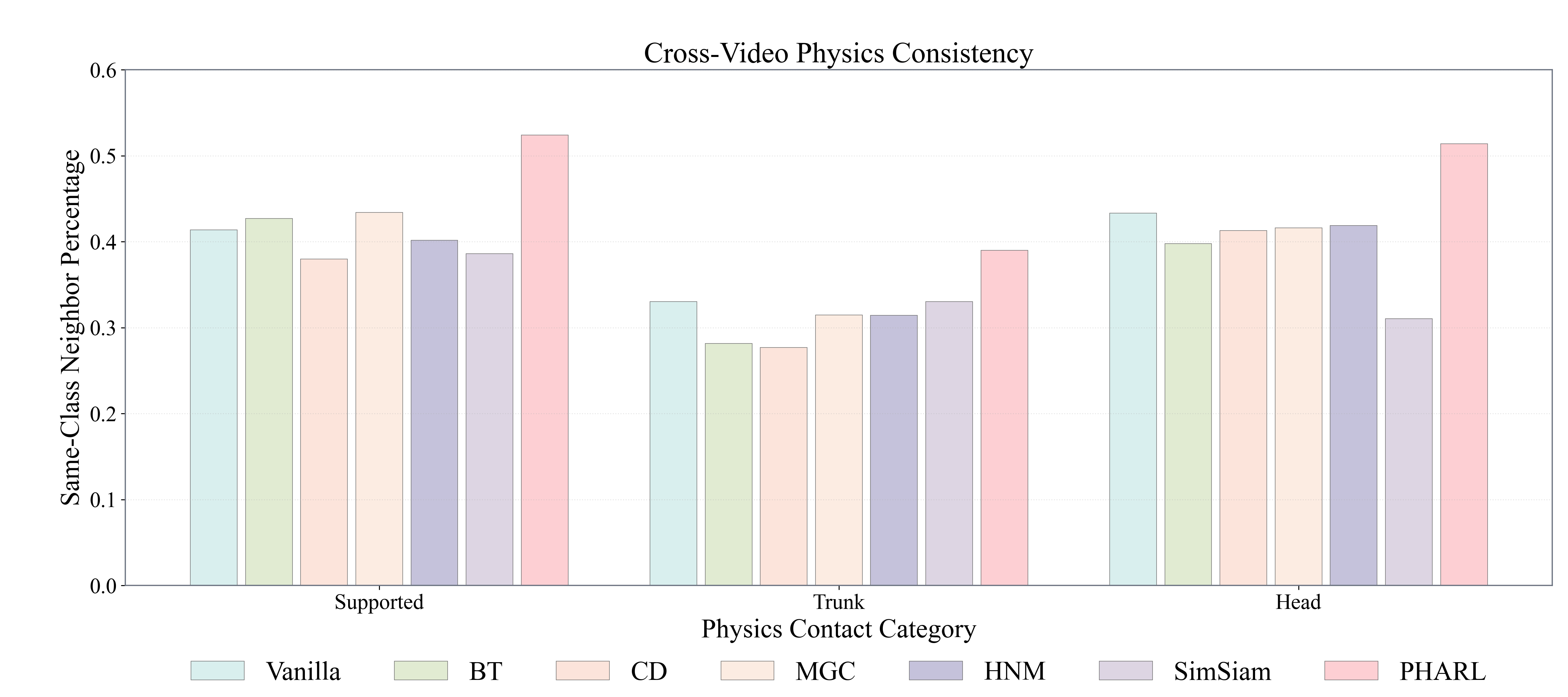}
	\end{center}
	\caption{
		Cross-video neighborhood consistency across physics contact categories. For each split, a class-balanced retrieval database is built, queries include all windows per class, and cosine nearest neighbors are retrieved after excluding same-video samples. Bars report row-normalized diagonal consistency (\textit{Supported@Supported}, \textit{Trunk@Trunk}, \textit{Head@Head}) with $k=10$. PHARL consistently achieves higher same-class consistency, indicating stronger cross-video physical structure in the learned representation.
	}
	\label{fig:neighborhood}
\end{figure}

Finally, we assess whether embeddings capture physical identity beyond individual video instances via cross-video neighborhood analysis (Figure~\ref{fig:neighborhood}). 
For each split, we build a class-balanced retrieval database by subsampling \textit{Supported}, \textit{Trunk}, and \textit{Head} windows to a common size. 
Queries are drawn from all windows in each class, and nearest neighbors are retrieved using cosine similarity in embedding space. 
To reduce video-specific shortcuts, same-video neighbors are excluded before selecting up to top-$k$ valid neighbors ($k=10$). 
PHARL yields higher diagonal consistency across impact categories than baseline methods. 
This suggests that the representation captures generalized physical characteristics associated with impact outcomes, rather than relying mainly on appearance cues or video-level context. 
Taken together, these analyses indicate that PHARL learns a latent geometry aligned with both ordinal structure and physical identity of impact outcomes.

\section{Discussion}\label{sec:discussion}

\subsection{Physics-Regularized Representations and Interpretation}
PHARL learns a representation geometry that is organized by physics-consistent outcomes (e.g., contact patterns and motion-level equivalences), rather than by visual or kinematic similarity alone. 
This behavior indicates that physics consistency guides the encoder toward latent factors that are relevant to contact dynamics, without introducing an explicit outcome-prediction objective. 
As a result, the learned embedding is useful for downstream analysis tasks such as motion retrieval, low-shot transfer, and expert-assisted review, while intentionally avoiding deployable severity scores or decision outputs.

\textbf{Weak Supervision vs. Self-Supervision.}
Our formulation is best interpreted as physics-guided weak supervision: simulation provides objective, scalable structural signals that shape embedding geometry without requiring manual injury labels. This setting is particularly attractive in safety-critical domains, where dense labels are expensive, inconsistent, or intrinsically difficult to define.

\textbf{Head-Contact Rarity and Its Implications.}
Head-contact windows remain a minority class (roughly 10\%, depending on split), which increases uncertainty in head-related comparisons and can visually compress class separation in 2D projections because of undersampling. 
Accordingly, we prioritize ordinality metrics (Spearman/POA), report contact AP/AUC as safety-localization utility, treat fall AUC as a deployment sanity check, and interpret PCR/Kendall together with qualitative plots as diagnostic evidence rather than deployable prediction evidence.

\subsection{Limitations and Future Work}
\textbf{Objective-level ordinal structure.} 
PHARL does not explicitly optimize an ordinal ranking loss.
Its physics term is defined through contact equivalence and binary grouping, so any \textit{Supported–Trunk–Head} ordering emerges post-hoc and is evaluated descriptively.
Future work may introduce explicit ordinal constraints (e.g., margin-based ranking losses) to test whether monotonic structure can be further strengthened without reframing the framework as an outcome or risk predictor.

\textbf{Simulation fidelity and environment factors.} 
The framework relies on a physics simulator that necessarily abstracts real-world fall dynamics.
Real events involve richer environment interactions (e.g., compliant or sloped surfaces, furniture contact, protective equipment), so the current objective emphasizes relative pattern extraction rather than absolute biomechanical fidelity.
Simulation hyperparameters (e.g., friction, restitution, contact thresholds) can influence specific $y^{\text{phys}}$ assignments.
Future work could include systematic sensitivity analysis and environment-aware simulation/modeling for improved real-world transfer, as well as more explicit treatment of protective behaviors beyond contact outcomes.

\textbf{Dataset bias and generalization.}
The current framework does not explicitly model dataset-specific biases (e.g., viewpoint, subject demographics), and cross-dataset generalization remains an open question.
Future work should incorporate bias-aware evaluation protocols and explicit domain-shift analyses across datasets and capture conditions.

\textbf{Initialization and motion decomposition.} 
Zero-velocity initialization improves numerical stability but may attenuate certain dynamic effects and reduce physical realism.
Future work could explore more physically grounded initialization strategies and improved decomposition of voluntary protective responses versus involuntary motion components to better balance stability and realism.

\subsection{Ethical Considerations and Deployment Constraints}
PHARL prioritizes scientific analysis and representation learning over direct severity prediction. 
At inference time, only RGB input is required; pose sequences, physics outcomes, and related kinematic metadata are used offline during training and evaluation only. 
Physics simulation is conducted on anonymized, non-identifiable data. 
The framework is intended for research and analytical use, not for clinical or safety-critical decision-making.

\section{Conclusion}\label{sec:conclusion}
We presented PHARL, a physics-regularized representation learning framework for fall-motion analysis without explicit injury annotations. 
PHARL addresses a central limitation of purely data-driven pipelines: motion-level similarity does not necessarily imply physics-consistent outcomes. 
By using simulation to construct outcome-level equivalence relations (instead of predicting outcomes directly), 
PHARL injects physical inductive bias as a structural regularizer for contrastive learning.
PHARL offers three practical benefits: 
(1) it avoids costly injury annotation, 
(2) it confines simulation to training while preserving efficient feed-forward inference, and 
(3) it provides a principled mechanism for incorporating domain knowledge through physics-derived structure. 
Experiments on four public datasets show consistent gains in contact-aware representation quality over motion-only baselines under linear-probe and geometry diagnostics.
Future work may extend this framework to other contact-rich activities, incorporate environment-specific physical interactions beyond ground contact, and explore multimodal extensions.

\bibliographystyle{Frontiers-Harvard}
\bibliography{src/reference}

@article{CAUCAFall,
  title={Dataset for human fall recognition in an uncontrolled environment},
  author={Guerrero, Jos{\'e} Camilo Eraso and Espa{\~n}a, Elena Mu{\~n}oz and A{\~n}asco, Mariela Mu{\~n}oz and Lopera, Jes{\'e}s Emilio Pinto},
  journal={Data in brief},
  volume={45},
  pages={108610},
  year={2022},
  publisher={Elsevier}
}

@article{GMDCSA-24,
  title={GMDCSA-24: a dataset for human fall detection in videos},
  author={Alam, Ekram and Sufian, Abu and Dutta, Paramartha and Leo, Marco and Hameed, Ibrahim A},
  journal={Data in Brief},
  volume={57},
  pages={110892},
  year={2024},
  publisher={Elsevier}
}

@article{Le2i,
  title={Optimized spatio-temporal descriptors for real-time fall detection: comparison of support vector machine and Adaboost-based classification},
  author={Charfi, Imen and Miteran, Johel and Dubois, Julien and Atri, Mohamed and Tourki, Rached},
  journal={Journal of Electronic Imaging},
  volume={22},
  number={4},
  pages={041106--041106},
  year={2013},
  publisher={Society of Photo-Optical Instrumentation Engineers}
}

@article{MCFD,
  title={Multiple cameras fall dataset},
  author={Auvinet, Edouard and Rougier, Caroline and Meunier, Jean and St-Arnaud, Alain and Rousseau, Jacqueline},
  journal={DIRO-Universit{\'e} de Montr{\'e}al, Tech. Rep},
  volume={1350},
  pages={24},
  year={2010}
}

@article{URFD,
  title={Human fall detection on embedded platform using depth maps and wireless accelerometer},
  author={Kwolek, Bogdan and Kepski, Michal},
  journal={Computer methods and programs in biomedicine},
  volume={117},
  number={3},
  pages={489--501},
  year={2014},
  publisher={Elsevier}
}

@inproceedings{shen2024world,
  title={World-grounded human motion recovery via gravity-view coordinates},
  author={Shen, Zehong and Pi, Huaijin and Xia, Yan and Cen, Zhi and Peng, Sida and Hu, Zechen and Bao, Hujun and Hu, Ruizhen and Zhou, Xiaowei},
  booktitle={SIGGRAPH Asia 2024 Conference Papers},
  pages={1--11},
  year={2024}
}

@inproceedings{chen2020simple,
  title={A simple framework for contrastive learning of visual representations},
  author={Chen, Ting and Kornblith, Simon and Norouzi, Mohammad and Hinton, Geoffrey},
  booktitle={International Conference on Machine Learning},
  pages={1597--1607},
  year={2020},
  organization={PMLR}
}

@inproceedings{he2020momentum,
  title={Momentum contrast for unsupervised visual representation learning},
  author={He, Kaiming and Fan, Haoqi and Wu, Yuxin and Xie, Saining and Girshick, Ross},
  booktitle={Proceedings of the IEEE/CVF Conference on Computer Vision and Pattern Recognition},
  pages={9729--9738},
  year={2020}
}

@article{shen2025multi,
  title={Multi-grained contrast for data-efficient unsupervised representation learning},
  author={Shen, Chengchao and Chen, Jianzhong and Wang, Jianxin},
  journal={Pattern Recognition},
  volume={165},
  pages={111655},
  year={2025},
  publisher={Elsevier}
}

@article{zhang2023patch,
  title={Patch-level contrasting without patch correspondence for accurate and dense contrastive representation learning},
  author={Zhang, Shaofeng and Zhu, Feng and Zhao, Rui and Yan, Junchi},
  journal={arXiv preprint arXiv:2306.13337},
  year={2023}
}

@inproceedings{yuan2023physdiff,
  title={Physdiff: Physics-guided human motion diffusion model},
  author={Yuan, Ye and Song, Jiaming and Iqbal, Umar and Vahdat, Arash and Hilliges, Otmar},
  booktitle={Proceedings of the IEEE/CVF International Conference on Computer Vision},
  pages={16010--16021},
  year={2023}
}

@article{zhang2022pimnet,
  title={Pimnet: Physics-infused neural network for human motion prediction},
  author={Zhang, Zhibo and Zhu, Yanjun and Rai, Rahul and Doermann, David},
  journal={IEEE Robotics and Automation Letters},
  volume={7},
  number={4},
  pages={8949--8955},
  year={2022},
  publisher={IEEE}
}

@article{gutierrez2021comprehensive,
  title={Comprehensive review of vision-based fall detection systems},
  author={Guti{\'e}rrez, Jes{\'u}s and Rodr{\'\i}guez, V{\'\i}ctor and Martin, Sergio},
  journal={Sensors},
  volume={21},
  number={3},
  pages={947},
  year={2021},
  publisher={MDPI}
}

@article{benkaci2024vision,
  title={Vision-based human fall detection systems: A review},
  author={Benkaci, Asma and Sliman, Layth and Dellys, Hachemi Nabil},
  journal={Procedia Computer Science},
  volume={241},
  pages={203--211},
  year={2024},
  publisher={Elsevier}
}

@article{guo2024msclr,
  title={Multi-Skeleton Contrastive Learning for Skeleton-Based Action Recognition},
  author={Guo, Tian and Liu, Hong and others},
  journal={IEEE Transactions on Multimedia},
  year={2024}
}

@inproceedings{yang2024cml,
  title={Contrastive Mask Learning for Action Recognition},
  author={Yang, Li and Wang, Zhaohui},
  booktitle={Proceedings of the IEEE/CVF Conference on Computer Vision and Pattern Recognition},
  year={2024}
}

@article{moore2024enhancing,
  title={Enhancing fall risk assessment: instrumenting vision with deep learning during walks},
  author={Moore, Jason and Catena, Robert and Fournier, Lisa and Jamali, Pegah and McMeekin, Peter and Stuart, Samuel and Walker, Richard and Salisbury, Thomas and Godfrey, Alan},
  journal={Journal of NeuroEngineering and Rehabilitation},
  volume={21},
  number={1},
  pages={106},
  year={2024},
  publisher={Springer}
}

@book{WHO2021,
  title={Step safely: strategies for preventing and managing falls across the life-course},
  author={{World Health Organization}},
  year={2021},
  publisher={World Health Organization}
}

@article{turaga2008machine,
  title={Machine recognition of human activities: A survey},
  author={Turaga, Pavan and Chellappa, Rama and Subrahmanian, Venkatramana S and Udrea, Octavian},
  journal={IEEE Transactions on Circuits and Systems for Video Technology},
  volume={18},
  number={11},
  pages={1473--1488},
  year={2008},
  publisher={IEEE}
}

@article{zhang2013fall,
  title={Fall detection using a wearable sensor and a robot},
  author={Zhang, Tong and Wang, Jue and Liang, Yanchun},
  journal={Computers in Biology and Medicine},
  volume={43},
  number={10},
  pages={1455--1463},
  year={2013},
  publisher={Elsevier}
}

@article{wang2020fall,
  title={Fall detection with multi-sensor fusion: A survey},
  author={Wang, Xinyu and Ellul, Joshua and Azzopardi, George},
  journal={IEEE Sensors Journal},
  volume={20},
  number={22},
  pages={13232--13251},
  year={2020},
  publisher={IEEE}
}

@article{li2020fall,
  title={Fall detection based on key points of human-skeleton using openpose},
  author={Li, Chao and Li, Qian and Li, Boshen and Feng, Lei},
  journal={Symmetry},
  volume={12},
  number={5},
  pages={744},
  year={2020},
  publisher={MDPI}
}

@article{de2016optimized,
  title={Optimized spatio-temporal descriptors for real-time fall detection},
  author={de Miguel, K and Bruna, A},
  journal={Journal of Ambient Intelligence and Humanized Computing},
  year={2016}
}

@article{dave2021tclr,
  title={Tclr: Temporal contrastive learning for video representation},
  author={Dave, Ishan and Gupta, Rohit and Shah, Mubarak and Sukthankar, Rahul},
  journal={Computer Vision and Image Understanding},
  volume={219},
  pages={103406},
  year={2022},
  publisher={Elsevier}
}

@inproceedings{qian2021spatiotemporal,
  title={Spatiotemporal contrastive video representation learning},
  author={Qian, Rui and Meng, Tianjian and Gong, Boqing and Yang, Ming-Hsuan and Wang, Hwehee and Belongie, Serge and Cui, Yin},
  booktitle={Proceedings of the IEEE/CVF Conference on Computer Vision and Pattern Recognition},
  pages={6964--6974},
  year={2021}
}

@article{rempe2021humor,
  title={HuMoR: 3D Human Motion Model for Robust Pose Estimation},
  author={Rempe, Davis and Birdal, Tolga and Hertzmann, Aaron and Guibas, Leonidas J},
  journal={arXiv preprint arXiv:2105.04668},
  year={2021}
}

@inproceedings{xie2021physics,
  title={Physics-guided human motion estimation with contact constraints},
  author={Xie, Kevin and Wang, Tingwu and Iqbal, Umar and Guo, Yunrong and Fidler, Sanja and Shkurti, Florian},
  booktitle={2021 IEEE International Conference on Robotics and Automation (ICRA)},
  pages={10274--10280},
  year={2021},
  organization={IEEE}
}

@inproceedings{zbontar2021barlow,
  title={Barlow twins: Self-supervised learning via redundancy reduction},
  author={Zbontar, Jure and Jing, Li and Misra, Ishan and LeCun, Yann and Deny, St{\'e}phane},
  booktitle={International Conference on Machine Learning},
  pages={12310--12320},
  year={2021},
  organization={PMLR}
}

@article{ma2024stgcn,
  title={Real-Time Fall Detection using Learnable Edges in ST-GCNs},
  author={Ma, Y. and Others},
  journal={arXiv preprint},
  year={2024}
}

@article{chen2025unigcn,
  title={Unified Graph Convolutional Networks for Multi-Task Human Motion Analysis},
  author={Chen, L. and Others},
  journal={IEEE Transactions on Cybernetics},
  year={2025}
}

@article{liu2024transformer,
  title={Transformer-based Fall Detection for Elderly Care: A Comprehensive Study},
  author={Liu, S. and Others},
  journal={Sensors},
  year={2024}
}

@article{wang2024view,
  title={Challenges in View-Invariant Fall Detection: A Contrastive Learning Perspective},
  author={Wang, Z. and Others},
  journal={Pattern Recognition},
  year={2024}
}

@article{karniadakis2021physics,
  title={Physics-informed machine learning},
  author={Karniadakis, George Em and Kevrekidis, Ioannis G and Lu, Lu and Perdikaris, Paris and Wang, Sifan and Yang, Liu},
  journal={Nature Reviews Physics},
  volume={3},
  number={6},
  pages={422--440},
  year={2021}
}

@article{ecri2025,
  title={Top 10 Health Technology Hazards for 2025},
  author={{ECRI Institute}},
  journal={ECRI Reports},
  year={2025}
}

@article{london2024,
  title={The London Protocol 2024: Systems Analysis of Clinical Incidents},
  author={Taylor-Adams, Sally and Vincent, Charles},
  journal={Clinical Risk},
  year={2024}
}

@article{robbins2014impact,
  title={The impact of body position and fall direction on head impact severity in elderly falls},
  author={Robbins, S. and Others},
  journal={Journal of Biomechanics},
  volume={47},
  pages={25--30},
  year={2014}
}

@article{choi2014head,
  title={Head impact biomechanics in sideways falls},
  author={Choi, W.J. and Robinovitch, S.N.},
  journal={Journal of Biomechanics},
  year={2014}
}

@article{duan2024revisiting,
  title={Revisiting Skeleton-based Action Recognition},
  author={Duan, H. and Others},
  journal={International Journal of Computer Vision},
  year={2024}
}

@article{yan2024skeleton,
  title={Skeleton-Based Action Recognition via Transformer: A Comprehensive Survey},
  author={Yan, S. and Others},
  journal={ACM Computing Surveys},
  year={2024}
}

@article{su2024motionbert,
  title={MotionBERT: A Unified Perspective on Learning Human Motion Representations},
  author={Su, Z. and Others},
  journal={Proceedings of the IEEE/CVF International Conference on Computer Vision},
  year={2024}
}

@inproceedings{lin2024skeletonmae,
  title={SkeletonMAE: Masked Autoencoders for Skeleton Action Recognition},
  author={Lin, Y. and Others},
  booktitle={Proceedings of the IEEE/CVF International Conference on Computer Vision},
  year={2024}
}

@article{islam2024deep,
  title={Deep Learning for Fall Detection: A Comprehensive Survey},
  author={Islam, M.M. and Others},
  journal={IEEE Access},
  year={2024}
}

@article{ahmed2024unsupervised,
  title={Unsupervised Anomaly Detection for Privacy-Preserving Fall Detection},
  author={Ahmed, S. and Others},
  journal={IEEE Internet of Things Journal},
  year={2024}
}

@inproceedings{shah2024anubis,
  title={ANUBIS: A Benchmark for Skeletal Action Recognition},
  author={Shah, A. and Others},
  booktitle={Advances in Neural Information Processing Systems},
  year={2024}
}

@article{jiang2024fine,
  title={Fine-Grained Representation Learning via Multi-Level Contrastive Learning without Class Priors},
  author={Jiang, Houwang and Liu, Zhuxian and Liu, Guodong and Liu, Xiaolong and Zhan, Shihua},
  journal={arXiv preprint arXiv:2409.04867},
  year={2024}
}

@article{kalantidis2020hard,
  title={Hard negative mixing for contrastive learning},
  author={Kalantidis, Yannis and Sariyildiz, Mert Bulent and Pion, Noe and Weinzaepfel, Philippe and Larlus, Diane},
  journal={Advances in Neural Information Processing Systems},
  volume={33},
  pages={21798--21809},
  year={2020}
}

@article{oord2018representation,
  title={Representation learning with contrastive predictive coding},
  author={Oord, Aaron van den and Li, Yazhe and Vinyals, Oriol},
  journal={arXiv preprint arXiv:1807.03748},
  year={2018}
}

@inproceedings{chen2021exploring,
  title={Exploring simple siamese representation learning},
  author={Chen, Xinlei and He, Kaiming},
  booktitle={Proceedings of the IEEE/CVF Conference on Computer Vision and Pattern Recognition},
  pages={15750--15758},
  year={2021}
}

@article{amundsen2022fall,
  title={Fall risk assessment and visualization through gait analysis},
  author={Amundsen, Tanner and Rossman, Matthew and Ahmad, Ishfaq and Clark, Addison and Huber, Manfred},
  journal={Smart Health},
  volume={25},
  pages={100284},
  year={2022},
  publisher={Elsevier}
}

@article{wang2023gait,
  title={Gait characteristics related to fall risk in patients with cerebral small vessel disease},
  author={Wang, Yajing and Li, Yanna and Liu, Shoufeng and Liu, Peipei and Zhu, Zhizhong and Wu, Jialing},
  journal={Frontiers in Neurology},
  volume={14},
  pages={1166151},
  year={2023},
  publisher={Frontiers Media SA}
}
	
\end{document}